\documentclass[letterpaper, 10 pt, conference]{ieeeconf}  % Comment this line out if you need a4paper

\IEEEoverridecommandlockouts                              % This command is only needed if 
\usepackage{graphicx}
\usepackage{wrapfig}
\usepackage{subfig}
\usepackage{algpseudocode}
\usepackage{array}
\usepackage{url} 
\usepackage{siunitx}
\usepackage{amsmath,amssymb}
\usepackage{mathrsfs}

\newtheorem{myprop}{Proposition}

\usepackage{mathtools}
\usepackage{bm}
\newtheorem{definition}{Definition}[section]
\usepackage[linesnumbered,ruled,vlined]{algorithm2e}
\usepackage{algpseudocode}
\usepackage{amsfonts}
\usepackage{tabularx}
\usepackage{threeparttable}
\usepackage{multirow}
\usepackage{lipsum}
\usepackage{stfloats} 
\usepackage{caption}
\newcommand{\specialcell}[2][l]{%
  \begin{tabular}[#1]{@{}l@{}}#2\end{tabular}}

\title{\LARGE \bf
Multi-Agent Reinforcement Learning Guided by\\
Signal Temporal Logic Specifications
}

\author{Jiangwei Wang$^{1}$, Shuo Yang$^{2}$, Ziyan An$^{3}$, Songyang Han$^{1}$, Zhili Zhang$^{1}$,\\ Rahul Mangharam$^{2}$, Meiyi Ma$^{3}$, Fei Miao$^{1}$
\thanks{$^{1}$Jiangwei Wang, Songyang Han, Zhili Zhang and Fei Miao are with University of Connecticut
        {\tt\small \{jiangwei.wang, songyang.han, zhili.zhang, fei.miao\}@uconn.edu}}%
\thanks{$^{2}$Shuo Yang and Rahul Mangharam are with University of Pennsylvania
        {\tt\small \{yangs1, rahulm\}@seas.upenn.edu}}%
\thanks{$^{3}$Ziyan An and Meiyi Ma are with Vanderbilt University
        {\tt\small \{ziyan.an, meiyi.ma\}@vanderbilt.edu}}%
}

\begin{document}

\maketitle
\thispagestyle{empty}
\pagestyle{empty}

%%%%%%%%%%%%%%%%%%%%%%%%%%%%%%%%%%%%%%%%%%%%%%%%%%%%%%%%%%%%%%%%%%%%%%%%%%%%%%%%
\begin{abstract}
Reward design is a key component of deep reinforcement learning (DRL), yet some tasks and designer's objectives may be unnatural to define as a scalar cost function. Among the various techniques, formal methods integrated with DRL have garnered considerable attention due to their expressiveness and flexibility to define the reward and requirements for different states and actions of the agent. However, how to leverage Signal Temporal Logic (STL) to guide multi-agent reinforcement learning (MARL) reward design remains unexplored. Complex
interactions, heterogeneous goals and critical safety requirements in multi-agent systems make this problem even more challenging. In this paper, we propose a novel STL-guided multi-agent reinforcement learning framework. The STL requirements are designed to include both task specifications according to the objective of each agent and safety specifications, and the robustness values of the STL specifications are leveraged to generate rewards. We validate the advantages of our method through empirical studies. The experimental results demonstrate significant reward performance improvements compared to MARL without STL guidance, along with a remarkable increase in the overall safety rate of the multi-agent systems.

\end{abstract}

%%%%%%%%%%%%%%%%%%%%%%%%%%%%%%%%%%%%%%%%%%%%%%%%%%%%%%%%%%%%%%%%%%%%%%%%%%%%%%%%
\section{INTRODUCTION}

%Reinforcement learning (RL) and multi-agent reinforcement learning (MARL) have gained significant research interests in solving various sequential decision-making problems in recent years, especially with the rapid development of deep reinforcement learning (DRL). It has been widely adopted in various domains such as robotics, healthcare systems, autonomous driving, smart cities and others that involve the participation of more than single agent and naturally fall into the realm of MARL~\cite{zhang2021multi}. It is a key component of an MARL algorithm to define a reward function that maps each state and an action that can be taken in that state to some real-valued reward~\cite{silver2021reward}. However, designing a reward function that informally capture the desired behavior and objective of the agent considering the dynamic interactions among multi-agents remains challenging and an open question. In particular, for multi-agent systems with decision-making problems that involve both the interactions of the agents and the physical dynamic process of each individual agent, poorly designed reward functions can lead to undesired policies that are unable to accomplish the tasks, or worse, execute unsafe actions in safety-critical systems~\cite{lu2021decentralized, zhang2022spatial}.

Multi-agent reinforcement learning (MARL) have gained significant research interests in solving various sequential decision-making problems for multi-agent systems, especially with the rapid development of deep reinforcement learning (DRL). It is a key component of an MARL algorithm to define a reward function that maps each state and action to some real-valued reward~\cite{silver2021reward}. However, define or encode a scalar reward function according to the desired behavior and objectives considering the dynamic interactions and typically heterogeneous goals of a multi-agent system remains challenging. Moreover, for MARL that involves both the interactions of the agents and the physical dynamic process of each individual agent, poorly designed reward functions can lead to undesired policies that are unable to accomplish the tasks, or worse, execute unsafe actions in safety-critical systems~\cite{lu2021decentralized, zhang2022spatial}.

\begin{figure}[h]
    \centering 
\subfloat[]{\includegraphics[height=1in]{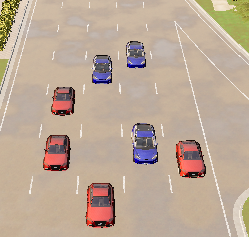}} \ \ 
\subfloat[]{\includegraphics[height=1 in]{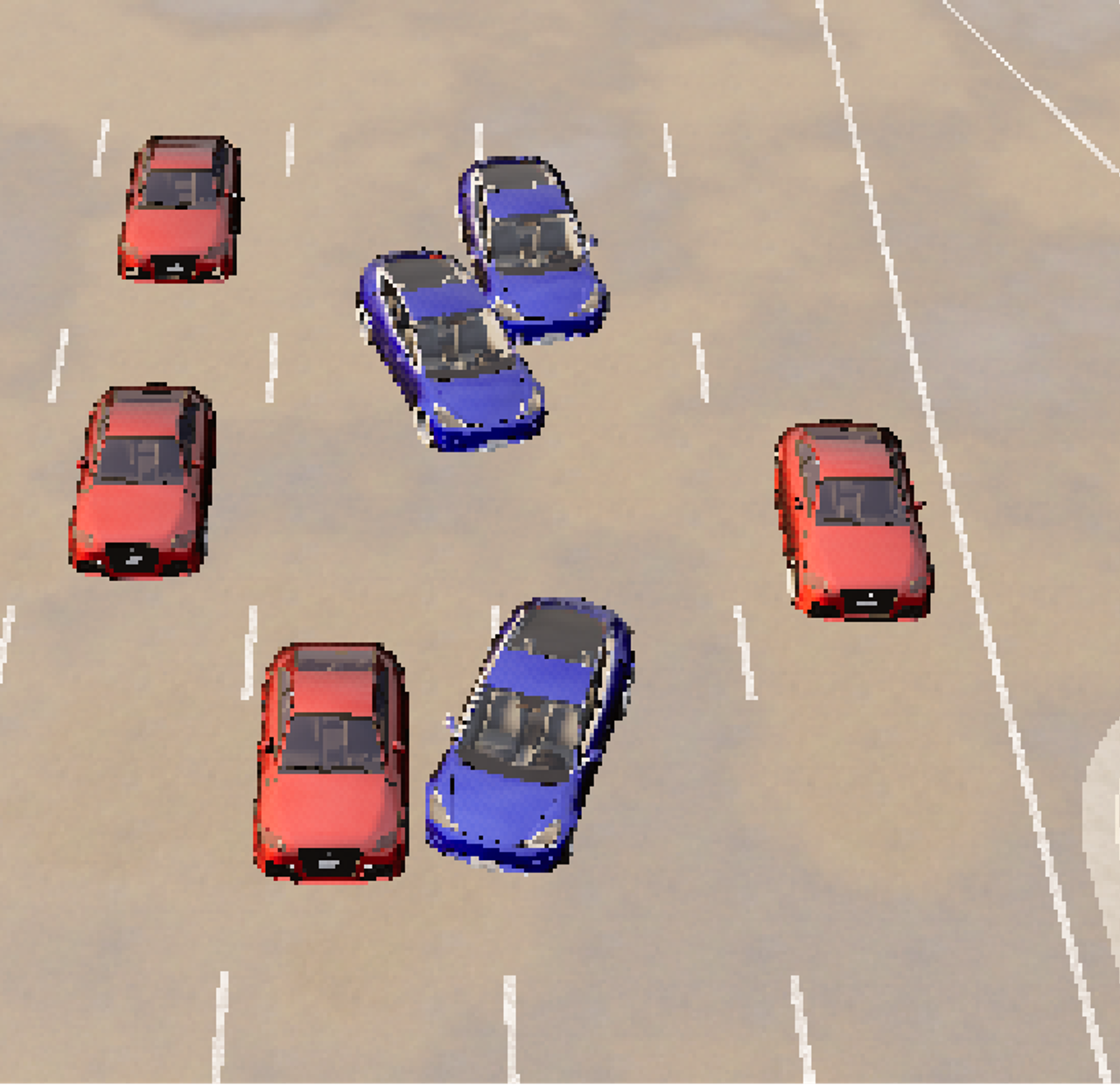}}  \ \ 
    \caption{\textit{Traffic-jam} in CARLA simulator. a: Scenario initialization. b: Agents trained based on MARL with hand-engineered reward fail to cross the open lane in a timely manner and collide with other agents.} 
    \label{fig:carla}
    \vspace*{-10pt}
\end{figure}

A \textit{motivating example} of decision-making for multi-agent system named \textit{Traffic-jam} is shown in Fig.~\ref{fig:carla}. In this case, red broken-down vehicles stopped on streets and blocked three lanes (due to an accident or other reasons), the three blue autonomous vehicles want to drive forward and pass through the only open lane (right lane in the figure) as soon as possible while keeping a safety distance between each other and the broken-down vehicles. Additionally, the time length which autonomous vehicles remain blocked after the broken-down vehicles should be constrained below a certain threshold. To make the vehicles learn to fulfill all these requirements, designing a reward function may take lots of trials and can be computationally expensive. Regarding the safety requirements in reinforcement learning \cite{garcia2015comprehensive, brunke2022safe, zhao2023state}, using penalty in reward to discourage the unsafe actions, or framing the problem into a constrained optimization problem may not provide sufficient safety assurances for the selected actions. Furthermore, it is even more challenging to encode a reward function considering temporal requirements for the agents~\cite{corazza2022reinforcement, kantaros2022accelerated}.

%To address this challenge, we propose a safe MARL algorithm that leverages signal temporal logic (STL) specifications and use their robustness values as the reward. 
To address this challenge, we propose a safe MARL algorithm that taking full advantages of signal temporal logic (STL), which, as a formal language  provides a more principled and expressive way to describe the requirements. We adopt the robustness values based on the STL requirements as rewards in our proposed STL-guided MARL algorithm. We also guarantee the satisfaction of hard safety requirements through an STL safety shield.
% The advantage of STL is that it can naturally capture the above-mentioned complex logical tasks with temporal requirements, and it has quantitative semantics to describe the \emph{satisfaction degree} of specifications. 
While there are extensive works exploring the use of temporal logic in single-agent RL, very little attention has been drawn on how to leverage signal temporal logic (STL) to guide MARL~\cite{balakrishnan2019structured, hammond2021multi}. 
%Overall, with the complex interactions and safety requirements in multi-agent systems, how to make the agents learn to fulfill the designer's goal and ensure the safety of the system remains open challenges. 
%In summary, our work addresses the challenges of reward shaping and safety requirements in MARL by leveraging STL specifications and providing additional safety guarantees.
%It has been shown that temporal logic languages can accelerate the training's convergence and reduce the reward hacking problem in single-agent reinforcement learning.
Our proposed algorithm shows promising results in learning a better policy for each agent to reach its objective and ensure the safety of the system. Our contributions are summarized as follows:
\begin{itemize}
    \item We design a multi-agent reinforcement learning algorithm that is guided by STL specifications, where the STL specifications include both safety requirements and the task that each agent aims to finish. We incorporate the STL safety shield to fulfill the safety specifications and provide safe actions for the agents.
\item The proposed algorithm utilizes STL to check partial trajectories and provide robustness values as a corresponding reward during the training process. 
\item We validate the proposed methodology in multi-agent particle-world environment (MPE) and CARLA testbeds. We demonstrate that compared with the baseline MARL algorithms and commonly used rewards, our proposed algorithm can learn better policies with both larger rewards and higher safety rates.
\end{itemize}
%present case studies on Multi-Agent Particle Environment (MPE) and 
\section{RELATED WORKS}
\subsection{Multi-Agent Reinforcement Learning}
There has been growing interest in the study of MARL since many real-world problems involve the interactions of multiple agents~\cite{zhang2021multi}. 
MARL approaches have been proposed for various multi-agent systems, such as unmanned aerial vehicles~\cite{cui2019multi, qie2019joint}, complex traffic networks~\cite{chu2019multi}, autonomous driving~\cite{shalev2016safe}, and so on.
Despite these successful applications, one remaining challenge in MARL is how to design good reward functions for complex tasks.
Poorly-designed reward functions might lead to undesired behavior and be detrimental to safety-critical systems~\cite{amodei2016concrete}.
The complex interactions among agents and their diverse objectives make the reward design hard.

\iffalse
Existing works that ensure system safety can be categorized into two types. The first type uses a model-based safety shield to ensure the safety of chosen actions. However, agents may fail to learn to choose safe actions and instead rely on the safety layer. The other type formulates safety requirements as constraints and trains agents to choose safe actions within the constraint. However, this method lacks guarantees for the actions chosen; agents may violate the safety constraint during training and execution. Additionally, the safety requirements in both of these types are relatively simple, such as maintaining a safe distance from other agents. More complex requirements, such as temporal requirements that mandate an agent be in a specific place at a given time, are difficult to achieve.
\fi
% \cite{lowe2017multi} proposes a multi-agent policy gradient algorithm where agents learn a centralized critic based on the observations and actions of all agents. \cite{lu2021decentralized}  considers that a team of agents cooperate in a shared environment, where each agent has its individual reward function and safety constraints that involve all agents' joint actions. As such, the agents aim to maximize the team-average long-term return, subject to all the safety constraints.
% (other MARL works? e.g. ~\cite{elsayed2021safe, zhang2019mamps, cai2021safe, zhang2022spatial, han2020multi, silver2021reward})
% \textcolor{blue}{first mention real systems examples (UAV, transportation, drones, robotic), then focus on MARL reward shaping difficulty;}
\subsection{Temporal Logic for Reinforcement Learning}
RL reward function usually relies on hand-engineered design or approaches like reward shaping~\cite{proper2012modeling}. In recent years, temporal logic specifications have been used extensively as training guidance in the context of single agent reinforcement learning for its power of expressiveness. 
In one direction, finite state automaton (FSA) is constructed to reward the agent (e.g., \cite{icarte2018using, li2019formal, cai2023safe})
%For example, authors of~\cite{li2019formal} convert the high-level linear temporal logic (LTL) specifications to the equivalent FSA, which is then used to guide the policy generation.
with the benefit of easy reward-generating automaton and high interpretability.
In another direction, the quantitative semantics of temporal logic formulas are captured to guide the policy training~\cite{balakrishnan2019structured, li2017reinforcement}.
%For example, \cite{balakrishnan2019structured} considers the robustness value of the given signal temporal logic requirement as the reward.
Simply applying the STL-guided single agent reinforcement learning in multi-agent setting is not a good solution because they don't consider the complex interactions between the agents and their safety requirements, which is usually the case in real world systems.

% \cite{icarte2018using} proposes a finite state machine-based reward machine combined with temporal logic to decompose the task and provides reward for different tasks. 
% In \cite{li2017reinforcement}, robustness value of the whole trajectory given the linear temporal logic requirement is used as the terminal reward to guide the reinforcement learning.  \cite{balakrishnan2019structured} then proposes to use signal temporal logic to check the partial trajectory during learning, then provide the robustness value as reward function for double DQN and PPO algorithm. A temporal logic based reward shaping is devised to provide additional reward through the learning process for continuous learning tasks \cite{jiang2021temporal}. 

In the context of MARL, very few works have been done to satisfy temporal logic specifications~\cite{hammond2021multi, elsayed2021safe,leon2020extended}. For example, \cite{hammond2021multi} proposes the first MARL algorithm for temporal logic specifications with correctness and convergence guarantees. 
However, it uses the LTL specifically designed to satisfy the non-Markovian, infinite-horizon specifications, which may not be applicable in the real world. Also, it has not been empirically verified in MARL environments. ~\cite{leon2020extended} proposes an extended Markov Games as a general mathematical model that allows multiple RL agents to concurrently learn various LTL specifications.
In our work, we consider STL specifications and use their robustness values as the rewards instead.
Compared with LTL, STL preserves quantitative semantics that can be used to establish a robust satisfaction value to quantify how well a trajectory fulfills a specification. This can further allow us to quantify the rewards more precisely and less sparser compared with LTL based approaches.

\section{Preliminary and Problem Formulation}
\subsection{Signal Temporal Logic}
In this section, we introduce the syntax, semantics, and robustness metric of STL, which is a powerful formal symbolism for specifying temporal logical requirements~\cite{maler2004monitoring}.
We first define a signal $\omega = \bm{s}^0\bm{s}^1\cdots \bm{s}^{t}$ as the state trajectory from starting state to time point $t$.
\begin{definition}
The syntax of STL is defined by: \[ \varphi ::= 
\mu
\mid \neg \varphi
\mid \varphi_1 \wedge \varphi_2
\mid \varphi_1 \vee \varphi_2
\mid \Diamond_{I} \varphi
\mid \square_{I} \varphi
\mid \varphi_1 \mathcal{U}_{I} \varphi_2 \]
\end{definition}

where $I=[a,b], \; a,b \in \mathbb{R}^{\geq 0}$ denotes a bounded time interval. The atomic predicate $\mu$ represents the underlying function $\mu(\omega^t) \geq 0$, where $\omega^t$ is the signal value at time $t$. 
We use the symbols $\square$, $\Diamond$, and $\mathcal{U}$ to denote temporal operators \textit{always}, \textit{eventually}, and \textit{until}. 
% which indicates that a requirement must be met at all times in the future. Similarly, we use  to denote the \textit{eventually} temporal operator, which requires that satisfaction must occur at some point in the future. Finally, we use $\mathcal{U}$ to denote the \textit{until} temporal operator, which specifies $\varphi_1$ is valid until $\varphi_2$ is true. 
The satisfaction relation $(\omega, t) \models \varphi$ evaluates to true ($\top$) if the specification $\varphi$ is satisfied by $\omega$ starting from $t$ and false ($\bot$) otherwise. 
In addition to its Boolean semantics, STL also owns the quantitative semantics (i.e. \emph{robust satisfaction values}), which quantify the degree of satisfaction~\cite{maler2004monitoring, fainekos2009robustness, deshmukh2017robust}.
The quantitative semantics assign real-valued measurements to the satisfaction (positive values) or violation (negative values) of the STL formula. 
In the evaluation section of this work, we utilize the robustness values to measure specification satisfactions.
\begin{definition}
STL quantitative semantics are defined in Table~\ref{table:stl-semantics}.
\vspace{-0.25cm}
\begin{table}[h!]
    \caption{STL quantitative semantics}
    \label{table:stl-semantics}
    \centering
    \begin{tabular}{p{2.5cm} p{4.5cm}}
        \hline
        Formula & Semantics \\
        \hline 
        $\rho(x \sim c, \varphi, t)$ &  $f(\omega[t] ) - c$  \\
        $\rho(\neg \varphi, \omega, t)$ & $- \rho(\varphi, \omega, t)$ \\
        $\rho(\varphi_1 \land \varphi_2, \omega, t)$ & $\min\{\rho(\varphi_1, \omega, t), \rho(\varphi_2, \omega, t) \}$ \\
        $\rho(\square_I \varphi, \omega, t)$ & $\underset{t' \in (t, t+I)}{\min} \rho(\varphi, \omega, t')$ \\
        $\rho(\Diamond_I \varphi , \omega, t)$ & $\underset{t' \in (t, t+I)}{\max} \rho(\varphi, \omega, t')$ \\
        $\rho(\varphi_1 \mathcal{U}_I \varphi_2, \omega, t)$ & \specialcell{$\sup_{t'\in (t + I) \cap \mathbb{T}} (\min\{\rho(\varphi_2, \omega, t'),$\\ $\inf_{t''\in[t,t']}(\rho(\varphi_1, \omega, t'')) \})$ } \\
        \hline
    \end{tabular}
\end{table}

% \begin{align*}
% &\rho(x \sim c, \varphi, t) &&= \pi(\omega[t] ) - c \\
% &\rho(\neg \varphi, \omega, t) &&= - \rho(\varphi, \omega, t) \\
% &\rho(\varphi_1 \land \varphi_2, \omega, t) &&=  \min\{\rho(\varphi_1, \omega, t), \rho(\varphi_2, \omega, t) \}\\
% % &\rho(\varphi_1 \vee \varphi_2, \omega, t) &&=  \max\{\rho(\varphi_1, \omega, t), \rho(\varphi_2, \omega, t) \}\\
% & \rho(\square_I \varphi, \omega, t) && = \underset{t' \in (t, t+I)}{\min} \rho(\varphi, \omega, t') \\
% & \rho(\Diamond_I \varphi , \omega, t) && = \underset{t' \in (t, t+I)}{\max} \rho(\varphi, \omega, t')\\
% &\rho(\varphi_1 \mathcal{U}_I \varphi_2, \omega, t) &&= \sup_{t'\in (t + I) \cap \mathbb{T}} (\min\{\rho(\varphi_2, \omega, t'),  \\
% & && \quad \, \inf_{t''\in[t,t']}(\rho(\varphi_1, \omega, t'')) \}) \\
% \end{align*}
\end{definition}
\vspace{-0.35cm}

\subsection{Problem Formulation of STL-guided MARL}

\textbf{STL-guided MARL}: In this work, we define an STL-guided MARL for multi-agent decision-making problem such as the example shown in Fig.~\ref{fig:carla}, to address the challenge of designing a reward function that utilizes the strength of STL. In particular, we define a tuple $G = (\mathcal{S}, \mathcal{A}, \mathcal{O}, \mathcal{T}, \{r_i\}, \gamma, \{\varphi_i\}, \omega^{t-L+1: t})$, where $\mathcal{S}$ is the joint state space, $\mathcal{A}$ is the joint action space, $o_i = \mathcal{O}(\bm{s};i)$ is the local observation for agent $i$ at global state
$\bm{s}$.  $\mathcal{T}$ corresponds to the state transition function as defined for Markov games in the literature ~\cite{littman1994markov}. The key new components of the tuple definition include: $\omega^{t-L+1: t} = \bm{s}^{t-L+1}\cdots \bm{s}^{t}$\footnote{To increase readability, we will omit truncation time indices $(t-L+1: t)$, i.e., we will use $\omega$ instead of $\omega^{t-L+1: t}$ to denote the partial trajectory with a slight notation abuse.} being the partial state trajectory of length $L$, and $\varphi_i$ being the STL formula for agent $i$.
Take the \textit{Traffic-jam} scenario as an example, the STL formula $\varphi_i$ is the aggregation of several STL requirements, including reaching the destination, keeping safe distance to other agents, and waiting no more than $T_{max}$ time steps after the broken-down vehicles. The detail of the STL formula for different tasks will be introduced in Section~\ref{section:experiment}.
In the tuple $G$, reward $r_i = \rho(\varphi_i, \omega, t)$ represents the STL robustness value, and each agent $i$ aims to maximize its own total expected return $R_i = \sum_{t=0}^{T} \gamma^t r_i^t$ where $\gamma$ is a discount factor and $T$ is the time horizon.

\section{Methodology}

\begin{figure}[h]
    \centering
    \includegraphics[height=1.7in]{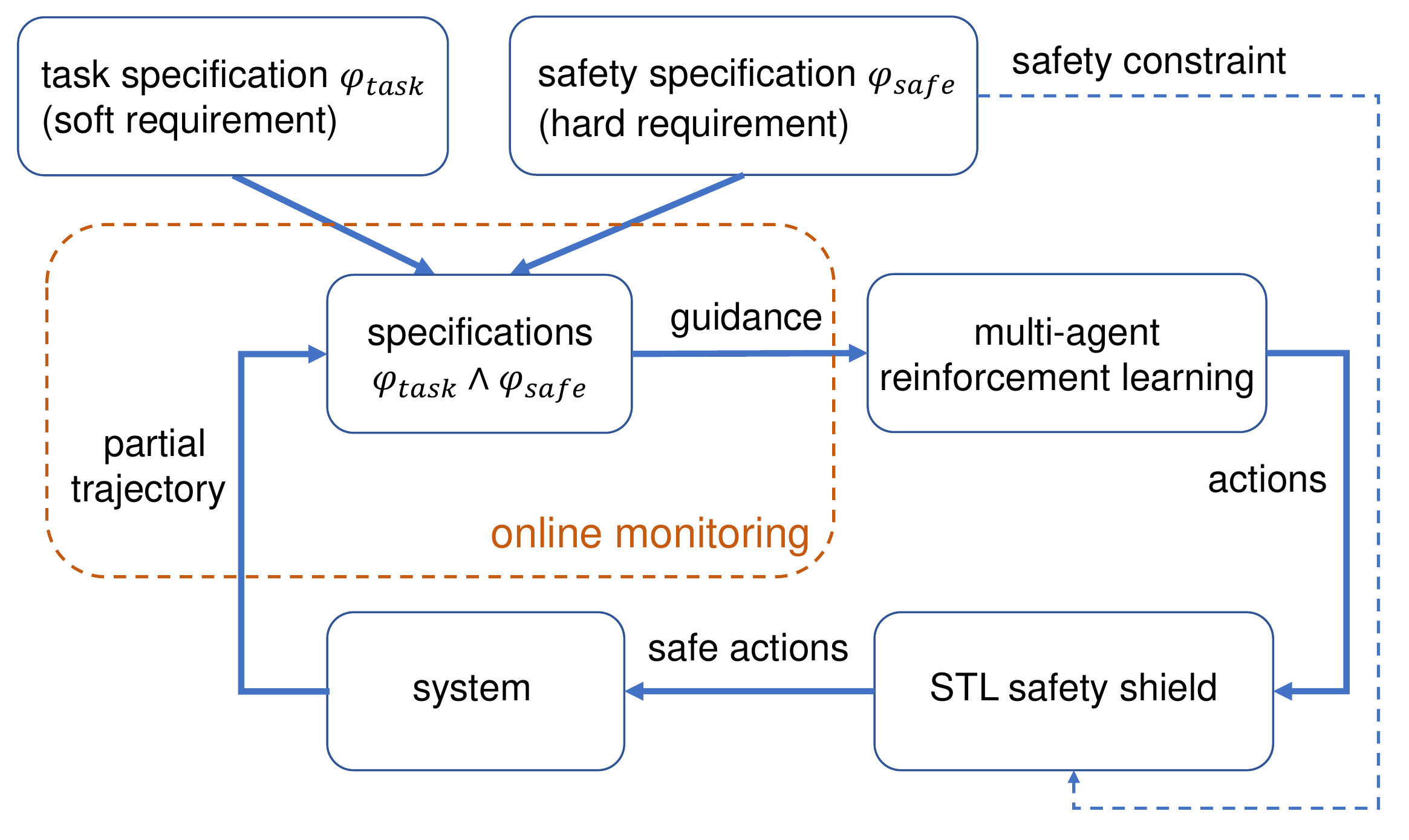}
    \vspace{-8pt}
    \caption{Methodology overview. The user-provided task specification and safety specification are expressed in STL formula $\varphi_{task}$ and $\varphi_{safe}$, respectively. The robustness values of the (partial) trajectory/signal w.r.t. $\varphi_{task}$ and $\varphi_{safe}$ are used to generate reward and guide the MARL policy learning. The STL safety shield, which is constructed based on safety specification $\varphi_{safe}$, is involved to safeguard the decisions made by MARL.}
    \label{fig:framework}
    \vspace*{-10pt}
\end{figure}

In this section, we first introduce the algorithm structure of STL-guided MARL algorithm. 
%The major novelties of the algorithm are: (1) we leverage the quantitative semantics of STL specifications including both task specifications and safety specifications; (2) during the training process, the robustness values of partial trajectories of each agent are generated by checking the STL specifications and used as rewards; (3) we introduce the CBF safety shield that fulfills the safety specifications and provides safe actions for the agents.
Our framework is shown in Fig.~\ref{fig:framework}.

\subsection{Reward Function}
To address the challenge of defining a reward function for a multi-agent system that considers the objective of each individual agent and the complex interactions among agents, in this work, we check the partial trace based on the STL specifications, and provide the robustness value as the reward. It's notable that our method is generalizable to different MARL algorithms, e.g., MADDPG~\cite{lowe2017multi} and MAPPO~\cite{yu2022surprising}.

% A partial trajectory in the multi-agent system starting from time $\tau$ to $\tau+L-1$ is defined by $\omega^{\tau} = (\textbf{x}^{\tau}, \dots, \textbf{x}^{\tau+L-1})$, where $\textbf{x}^t = (o_1^t, \dots, o_N^t)$ is the global state. 
At each time step of the training process, the trajectory contains the states of all the agents. Given the partial trajectory and STL specification $\varphi_i$, which might consist of both safety specification $\varphi_{i, safe}$ and task specification $\varphi_{i, task}$, the reward for agent $i$ can be defined as
\begin{equation}
    r^t_i = \rho(\varphi_i, \omega, t), \forall t \in [\tau, \tau+L-1],
\end{equation}
where $\rho$ is the robustness value. % that is defined in Appendix~\ref{appendix:stl-quantitative}.
Given the partial trajectory, the more it satisfies the STL specifications, the larger reward it obtains.
One reason we use partial trajectory robustness value rather full trajectory is that the former better discriminates different states and actions, while it's hard for latter to assign the credit to specific actions, and estimating the value function also becomes more challenging with longer future horizons. From MARL algorithm design perspective, n-step methods allow for credit assignment over a longer time horizon such as in the A3C algorithm \cite{mnih2016asynchronous}. This enables the agent to better understand the consequences of its actions. Hence, in this work, we define the reward based on the robustness value given the partial trajectory $\omega$.
It should also be noted that the STL requirements encompass various types, including reaching the goal, safety requirements like maintaining a safe distance from other agents, and other temporal requirements. Detailed STL requirements for task specification and safety specification are illustrated by a case study in Subsection~\ref{case_study}.

\vspace*{-10pt}
\begin{algorithm}[!h]
    \SetAlgoLined
    \LinesNumbered
    \caption{Pseudocode for STL-guided MARL}
    \label{alg:one}
    
    \SetKwInOut{Input}{Input}
    \SetKwInOut{Output}{Output}
    
    %\Input{Input parameters}
    %\Output{Output result}
    
    Orthogonal initialization for $\theta_i$ and $\phi_i$, the parameters for policy $\pi_i$ and critic $V_i$, respectively\;
    
    \For{episode = 1 to M}{
        Episode initialization: replay buffer $\mathcal{D}\leftarrow \emptyset$; initial state $\bm{s}$; step $t \gets 1$; rollout step number $L$\;
        %Initialize 
        
        \While{$t \leq t_{\max}$}{
        %\scriptsize 
            $t_{\text{start}} = t$\;
            Initialize an empty trajectory $\omega$\;
            
            \For{t = $t_{\text{start}}$ to $t_{\text{start}}+L-1$}{
                For each agent $i$, select action $a_{i_q}^t \gets \pi_{\theta_i}(o_i^t)$, send $a_{i_q}^t$ into STL safety shield, and send back $a_i^t$\;
                
                Execute actions $\bm{a}^t = (a_1^t, \dots, a_N^t)$ chosen based on STL safety shield, append $\bm{s}^t$ to $\omega$, observe reward $r^t_i = \rho(\varphi_i, \omega, t)$ for each agent $i$ given agent $i$'s STL formula $\varphi_i$, observe new state $\bm{s}^{t+1}$\;
                
                $\bm{s}^t \gets \bm{s}^{t+1}$; $t \gets t+1$\;
            }
        }
        %$(\textbf{x}^{t_{\text{start}}}, \bm{a}^{t_{\text{start}}}, \bm{r}^{t_{\text{start}}}, \textbf{x}^{t_{\text{start}}+1}, \dots, \textbf{x}^{t_{\text{start}}+L-1}, \bm{a}^{t_{\text{start}}+L-1}, \bm{r}^{t_{\text{start}}+L-1}, \textbf{x}^{t_{\text{start}}+L})$
        Store in $\mathcal{D}$\ $\{(\bm{s}^{\tau}, \bm{a}^{\tau}, \bm{r}^{\tau}, \bm{s}^{\tau+1})\ |\ \tau \in [t_{\text{start}},\dots,t_{\text{start}}+L-1]\}$;
        \For{agent $i = 1$ to N}{
        %\scriptsize
            % $(\textbf{x}^{t}, \bm{a}^{t}, \bm{r}^{t}, \textbf{x}^{t+1}, \dots, \textbf{x}^{t+L-1}, \bm{a}^{t+L-1}, \bm{r}^{t+L-1}, \textbf{x}^{t+L})$
            Randomly sample mini-batch from $\mathcal{D}$ $\{(\bm{s}^{\tau}, \bm{a}^{\tau}, \bm{r}^{\tau}, \bm{s}^{\tau+1})\ |\ \tau \in [t,\dots,t+L-1]\}$\;
            Update critic by minimizing $ \mathcal{L}^{\text{VF}}(\phi_i)$\;
            Update actor by maximizing $ \mathcal{L}(\theta_i)$\;
        }
    }
\end{algorithm}
\vspace*{-25pt} % Reduce space

\iffalse
\textbf{Discussion 1:}
  Note that the online robustness value can be efficiently computed using existing approaches, see, e.g.,~\cite{deshmukh2017robust}, and there are some toolboxes that can be easily leveraged such as \texttt{pcheck}\footnote{\url{https://github.com/simonesilvetti/pcheck}} and \texttt{stlcg}\footnote{\url{https://github.com/StanfordASL/stlcg}}.  
\fi
%In this work, we adopt the centralized training decentralized execution paradigm. As shown in Alg~\ref{alg:one}, during training, in roll out time steps $T$, each agent chooses action $a_{i_q}^t$ based on their local observations $o_i^t$, this action is then sent to the STL-CBF layer to make sure the action can ensure the safety and fulfill the STL requirements. The details of STL-CBF will be introduced in next section. Global state $\textbf{x}^t = (o_1^t, \dots, o_N^t)$, which is the aggregation of the local observations, is appended to the previous states to form the partial trajectory $\omega$, by checking the partial trajectory and the STL requirements, each agent gets the reward $r_i^t$. Note that the STL requirements can be summarized into different types, including reaching the goal, keeping safe distance to other agents, and other temporal requirements. The detailed STL requirements for different tasks are illustrated in Section~\ref{section:experiment}. 

%\textbf{todo}: can add discription of the network, use rnn to take the trajectory. 

\subsection{Algorithm Structure}

In our work, we adopt the centralized training and decentralized execution paradigm. As depicted in Alg.~\ref{alg:one}, during training, in rollout time steps $t$, each agent selects potentially unsafe action $a_{i_q}^t$ based on its local observations $o_i^t$. This action is then passed to the STL safety shield layer, and safe action $a_i^t$ is returned and deployed such that the satisfaction of the STL safety specifications is guaranteed. The details of the STL safety shield layer will be presented in the next section. The global state $\bm{s}^t = (o_1^t, \dots, o_N^t)$, representing the aggregation of the local observations, is appended to the previous states to form the partial trajectory $\omega$. By evaluating the partial trajectory against the STL requirements, each agent obtains its reward $r_i^t$.

The actor network is trained to maximize the following objective:
\vspace*{-6pt}
\begin{equation}
\begin{aligned}
   \mathcal{L}(\theta_i) & = \frac{1}{B} \sum_{k=1}^{B} \text{min}(r_{\theta_i, k}A_{i,k}, \text{clip}(r_{\theta, k}, 1-\epsilon, 1+\epsilon)A_{i,k}) \\
       & +\sigma \frac{1}{B} \sum_{k=1}^{B} S[\pi_{\theta_i}(o_{i,k})],
   \end{aligned}
\end{equation}
where $B$ is the batch size, $r_{\theta_{i, k}} = \frac{\pi_{\theta_i}(a_{i,k}|o_{i,k})}{\pi_{\theta_{i_{old}}}(a_{i,k}|o_{i,k})}$ denotes the ratio of the probability under the new and old policies respectively, $A_{i,k}$ is the advantage computed by GAE method \cite{schulman2015high}, $S$ is the policy entropy, and $\sigma$ is the entropy coefficient hyperparameter. 
For the critic network, the loss function is:
\vspace*{-10pt}
\begin{equation}
\begin{aligned}
     & \mathcal{L}^{\text{VF}}(\phi_i)  = \frac{1}{B} \sum_{k=1}^{B} \text{max}[(V_{\phi_{i}}(\bm{s}_{i,k}) - \hat{R}_{i,k})^2, \\ & (\text{clip}(V_{\phi_i}(\bm{s}_{i,k}), V_{\phi_{i_{\text{old}}}}(\bm{s}_{i,k})-\epsilon, V_{\phi_{i_{\text{old}}}}(\bm{s}_{i,k})+\epsilon) - \hat{R}_{i,k})^2],
\end{aligned}
\end{equation}

where $\hat{R}_{i,k}$ is the discounted reward-to-go. For actor and critic network, we use the recurrent neural network (RNN) to take the input to enable the agents to effectively model and reason about sequential information in their interactions with the environment. By maintaining hidden states and updating them at each time step, RNN can capture the temporal dynamics and dependencies across multiple time steps.

\subsection{STL Safety Shield}

In multi-agent systems with complex dynamics, such as the \textit{Traffic-jam} scenario depicted in Fig.~\ref{fig:carla}, ensuring system safety specification becomes paramount. One critical aspect of safety is maintaining a safe distance between agents. Consequently, we incorporate safety requirements into the Signal Temporal Logic (STL) formulas $\varphi_{safe}$, which will also be elaborated in Section~\ref{section:experiment}. To satisfy these safety requirements specified in the STL formulas, such as ``the distance between agents should always be greater than a threshold'', we first convert them to CBFs. 
Then, we employ the quadratic programming (CBF-QP) to ensure safety. By leveraging CBF-QP, we assess whether each discrete action guarantees system safety and filter out any unsafe actions accordingly.

%\paragraph{Control barrier functions}

\iffalse
\textbf{Discussion 2:}
Note that the low-level control state space is different from the action space. 
We adopt the widely used kinematic bicycle model for its simplicity while still considering the non-holonomic vehicle behaviors \cite{kong2015autonomous}.% (see Appendix~\ref{appendix:bicycle}).The state of the vehicle is $\bm{x}^t= [x^t,y^t, \psi^t, v^t]$, where $x^t$ and $y^t$ denote the coordinates of the vehicle's center of gravity (c.g.) in an inertial frame ($X,Y$), $\psi^t$ and $v^t$ represent the orientation and velocity of the vehicle. 
 Here the vehicle's state $\bm{x^t}$ is part of its observation $o_i^t$, whose definition is in Section~\ref{section: carla}. 
Also, the low-level control is different from Markov game action. The inputs $\bm{u}^t$ of the low-level system are the acceleration $a^t$ and the steering angle $\delta_f^t$. Given each discrete action $a_{i_q}^t$ of Markov game, e.g., keep lane, change to left or right lane, acceleration or deceleration, there will be a corresponding continuous control input $\bm{u}_{i_q}^t$ using the trajectory planning method \cite{dixit2018trajectory, cesari2017scenario}. \\
\textbf{Discussion 3: } Safe reinforcement learning using CBFs has also drawn some attention recently, see, e.g.,~\cite{cheng2019end, emam2022safe}. However, these works consider single-agent setting and time-invariant CBFs, but we consider multi-agent setting and thus time-varying CBFs instead.
\fi

\subsubsection{Case Study}
\label{case_study}
We use \textit{Traffic-jam} scenario to explain the details of STL safety shield design and the STL specifications. As shown in Fig.~\ref{fig:carla}, in the \textit{Traffic-jam} scenario, red broken-down vehicles block three lanes, a group of autonomous vehicles aim to cross the narrow road and arrive their destination as soon as possible, while maintaining the safety of the whole system. Agent $i$'s observation at time $t$ include (1) its own locations, velocities, accelerations, orientation $(\bm{p}_i^t, \bm{v}_i^t, \bm{a}_i^t, \psi_i^t)$; (2) other agent $j$'s shared information $(\bm{p}_j^t, \bm{v}_j^t, \bm{a}_j^t, \psi_j^t), \forall j \in N$, (3) its destination $\bm{p}_{i_{dest}}$. Agent $i$'s discrete action space include: $a_{i,1}$: keep speed and keep in current lane; $a_{i,2}$: change to left lane; $a_{i,3}$: change to right lane; $a_{i,4}$: brake; $a_{i,5}\sim a_{i,4+l}$: $l$ different throttle values, representing $l$ levels of acceleration and deceleration in the current lane.

We have the following requirements for each agent $i$:  
(1) (safety) distance between itself and the leading vehicle in its current lane and neighboring lanes should be always greater than a safe distance;
(2) (task) eventually reach its destination;  
(3) (task) it should stop in front of the narrow road location $\bm{p}_{road}$; 
(4) (task) its blocked duration $t_{wait}$ in front of narrow road location $\bm{p}_{road}$ should be less than $T_{\text{max}}$.
These specifications can be easily converted to STL formulas accordingly:
\vspace{-10pt} % Reduce space
\begin{equation}
\label{eq:trafficjam}
\begin{aligned}
    &\varphi_{i_1} = \Box_{[0, T-1]} \|\bm{p}^t_i - \bm{p}^t_{j}\| - \frac{(v^t_{j} - v^t_i)^2}{2 a_{i_l}} \ge \epsilon_1, \forall j \in N, j \neq i \\
    &
    \begin{cases}
      \varphi_{i_2} = \Diamond_{[0, T-1]} \|\bm{p}^t_i - \bm{p}_{i_{\text{dest}}}\| \le \epsilon_2\\
      \varphi_{i_3} = \Box_{[0, T-1]} \, (\neg (\|\bm{p}^t_i - \bm{p}_{\text{road}}\| \leq L) \vee (\Diamond_{[0, \tau]} v^t_{i} \leq 0))\\
      \varphi_{i_4} = \Box_{[0, T-1]} (\neg (\|\bm{p}^t_i - \bm{p}_{\text{road}}\| \le L) \vee (t_{\text{wait}} < T_{\text{max}}))
    \end{cases}     
\end{aligned}
\end{equation}

%\textbf{todo}: in algorithm 1, the t is superscript, here t is subscript, shoould make consistent?

\textit{CBF for safety}:
Here we show how to define the barrier functions to fulfill the safety requirements in STL specifications. Notably, there is exiting work summarizes how to design CBF given different STL predicates generally \cite{lindemann2018control}. We model the low-level agent dynamic as a nonlinear control affine  system:  $\bm{x}^{t+1}=f(\bm{x}^t)+g(\bm{x}^t) \bm{u}^t$,
where $\bm{x}^t\in \mathbb{R}^n$, $\bm{u}^t\in \mathcal{U}$ with $\mathcal{U}\subseteq \mathbb{R}^m$ denoting the set of permissible control inputs, $f$ is the nominal unactuated dynamics, and $g$ is the nominal actuated dynamics. We adopt the widely used kinematic bicycle model for its simplicity while still considering the non-holonomic vehicle behaviors \cite{kong2015autonomous}.

\begin{figure}[h]
    \vspace*{-10pt}
    \centering
    \includegraphics[scale = 0.6]{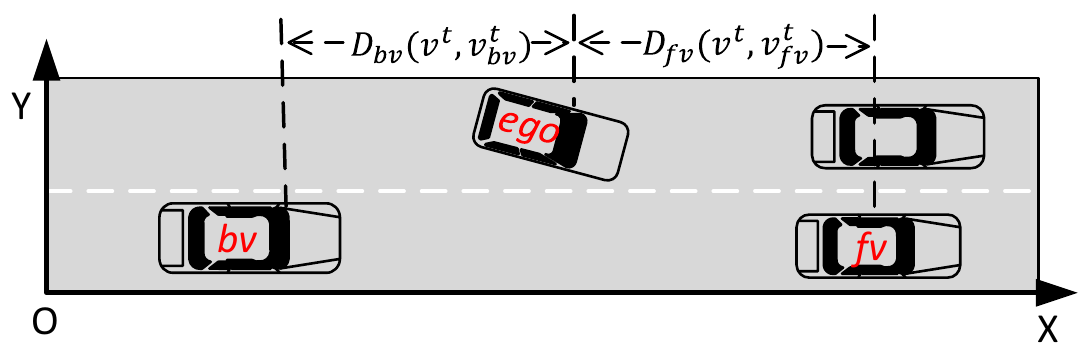}
    \caption{Lane change in \textit{Traffic-jam} scenario.}
    \label{fig:CBF}
    \vspace*{-20pt}
\end{figure}

As shown in Fig.~\ref{fig:CBF}, during the lane keeping mode, the ego vehicle should keep a safe distance to the vehicle in the front in its current lane; while it's changing the lane, it should keep a safe distance to both the vehicles in its front and back. 
We use $fv$ and $bv$ to denote the front and back vehicles in the target lane respectively. 
The safe distance to the front vehicle can be expressed as $D_{fv} = (1+\epsilon)v^t + \frac{(v^t-v_{fv}^t)^2}{2 a_l} $, and the safe distance to the back vehicle is $D_{bv} = (1+\epsilon)v^t + \frac{(v_{bv}^t-v^t)^2}{2 a_l} $. 
Note that $a_l$ and $v^t$ are the ego vehicle's acceleration limit and current speed respectively, and $v_{fv}^t, v_{bv}^t$ denote the velocity of vehicle in the front and back respectively. Finally, the CBFs can be expressed as:
$h_{fv}(\bm{x}^t, t) = (x_{fv}^t -x^t) - D_{fv}(v^t, v_{fv}^t) $ and $h_{bv}(\bm{x}^t, t) = (x^t -x^t_{bv}) - D_{bv}(v^t, v_{bv}^t)$.

For each discrete action $a_{i_q}^t$, after being mapped to the corresponding continuous control input $\bm{u}_{i_q}^t$\cite{dixit2018trajectory, cesari2017scenario}, then the following CBF-QP is solved to return a safe control input:
\begin{equation} \label{eq:safety_filter}
\begin{aligned}
      & {\text{min}} \quad \lVert \bm{u}^t - \bm{u}_{i_q}^t \rVert_2^2\\
     \text{s.t.} & \sup_{\bm{u}^t \in \mathcal{U}}[h(f(\bm{x}^t)+g(\bm{x}^t)\bm{u}^t, t+1)-h(\bm{x}^t, t)]\geq -\gamma h(\bm{x}^t, t).
\end{aligned}
\end{equation}
Then, we can have the safety guarantees~\cite{ames2016control, zeng2021safety}.
\begin{myprop}
\label{thm:cbf}
Assume $h(\bm{x}^t, t)$ is a valid time-varying CBF on $\mathcal{C}$. Then any controller $\bm{u}^t(\bm{x}^t)$ from~(\ref{eq:safety_filter}) for all $\bm{x}^t \in \mathcal{C}$ will render the set $\mathcal{C}$ forward invariant, i.e., the system is safe.
\end{myprop}

\section{Experiments and Case Studies}
\label{section:experiment}

\subsection{Testbeds and Common Experiment Setup}

\textbf{Testbed environment}: We evaluate the performance of STL-guided MARL algorithm on two benchmarks: the multi-agent particle-world environment (MPE) \cite{lowe2017multi} and CARLA \cite{dosovitskiy2017carla}. %In MPE testbed, we design two new scenarios that has more stages of destinations for the agent to reach, meaning more temporal requirements,  based on the existing tasks, we name then as simple coordination II and simple spread II. In CARLA testbed, we consider two scenarios, traffic jam and intersection, with complex interactions, harsh safety requirements, and more temporal requirements.
For MPE, we develop two new scenarios, namely ``Simple Coordination II'' and ``Simple Spread II'', which feature an increased number of stages for the agents to reach. These new scenarios introduce additional temporal requirements compared to the existing tasks. Within CARLA, we consider the \textit{Traffic-jam} and \textit{Traffic-jam Expansion} scenario, which are characterized by intricate interactions, demanding stringent safety requirements, and imposing higher temporal constraints on the agents' decision-making processes.

\textbf{Common experimental setup}: 
We conduct a comparative analysis of the STL-guided MARL algorithm and the MARL algorithm with the original task reward in both testbeds. In STL-guided MARL, the STL reward are the weighted sum of the robustness values of all the STL specifications.
We denote the STL reward as $ r_i^t = \sum_{j} c_j \rho(\varphi_{i_j}, \omega, t) + b$, where $c_j$ are weights and $b$ is a constant.
To ensure a fair comparison, the original reward function are based on widely adopted designs found in the existing literature. We evaluate the performance of both methods by examining the episode return. We compute the return using the STL reward in both algorithms for consistency. This approach allows us to quantify the extent to which the agents learn to fulfill the designer's goals. For our experiments, we utilize a server equipped with Intel Core i9-10900X processors and four NVIDIA RTX2080Ti GPUs. The experiments are conducted using Python 3.6.0, PyTorch 1.6.0, and CUDA 11.0. 
%In the following sections for the two testbeds, we first introduce the reward function used for baseline MARL algorithms, then introduce the STL specifications of our proposed Alg.1, finally we present our experiment results. The parameters for experiment are in appendix.

%In both testbeds, we compare STL-guided MARL alogrithm with the MARL algorithm with the original task reward. For a fair comparison, The original reward function are based on the widely adopted reward design in the existing literature. We then show the episode return for both methods. Note that in both algorithms, the return are computed using the STL reward, since we want to quantify to what extent does the agent learn to fulfill the designer's goals. The larger the return, the more the agent fulfill the STL requirements. In our experiments, we use a server configured with Intel Core i9-10900X processors and four NVIDIA RTX2080Ti GPUs. Our experiments are performed on Python 3.6.0, PyTorch 1.6.0, and CUDA 11.0.

\subsection{MPE Testbed}
\label{sec:mpe}
\paragraph{Environment}In MPE, we design two new tasks, simple coordination II and simple spread II to evaluate our algorithm. In both tasks, the observation of agent $i$ include the relative positions to other agents and the landmarks, the discrete actions space are: stay, left, right, up and down.

\textbf{Baselines:} In both tasks, $N$ agents need to first cover $N$ landmarks in the first stage, and then another $N$ landmarks in the second stage with least collisions. The difference is: in simple coordination II, agent and landmark are paired so agent only targets at its own corresponding landmark; for simple spread II, agents learn to infer the landmark they must cover, and move there while avoiding other agents. The reward function for simple coordination II is: $  r = - c_1 \sum_{i= 1}^{N} ( |\bm{p}_i-\bm{p}_{i, \text{landmark, goal}}|) + c_2 \sum_{i= 1}^{N} (|\bm{p}_i-\bm{p}_{i, \text{landmark, others}}|) $ where $\bm{p}_i$ is location of agent $i$, $\bm{p}_{i, \text{landmark, goal}}$ is location of the current goal landmark of agent $i$, $\bm{p}_{i, \text{landmark, goal}}$ is location of the other landmark.  The reward function for simple spread II is $  r = - c_1 \sum_{i= 1}^{N} (\min_{j \in N} |\bm{p}_j-\bm{p}_{i, \text{landmark, goal}}|) + c_2 \sum_{i= 1}^{N} (\min_{j \in N} |\bm{p}_j-\bm{p}_{i, \text{landmark, others}}|) $
where $\bm{p}_j$ is location of agent $j$, $\bm{p}_{i, \text{landmark, goal}}$ is location of the landmark $i$ in current goal stage, $\bm{p}_{i, \text{landmark, others}}$ is location of the other stage's landmark $i$. In both tasks, there will be a $-1$ penalty added on the current reward for a collision. The reward function is based on the reward designed in the existing works\cite{lowe2017multi, li2017reinforcement}. We use MADDPG\cite{lowe2017multi} as our baseline algorithm.

\textbf{STL-guided MARL}: For both tasks, given a whole trajectory of length $T$ of agent $i$, the requirements include:  %are $ \phi =(\phi_1 \mathcal{T} \phi_2) \land (\neg \phi_2 \mathcal{U} \phi_1) \land \phi_4 $: 
(1) All first part of $N$ landmarks are eventually visited by their corresponding agent;
(2) All second part of $N$ landmarks are eventually covered by their corresponding agent; 
 (3) no collision between agents nearby (the distance is always greater than the safety threshold); (4) The first three landmarks should be visited at least once before the second three landmarks are visited.

 Therefore, we write the specifications for simple coordination as:    
 \vspace{-10pt} % Reduce space
\begin{equation}
\begin{aligned} 
       \varphi_{i_1} &=\Diamond_{[0,T]} \bigwedge_{1,2, \dots}|\bm{p}_i^t-\bm{p}_{i, \text{landmark, first}}|\le \epsilon_1, \\  
       \varphi_{i_2} &= \Diamond \Box_{[0, T]} \bigwedge_{1,2, \dots} |\bm{p}_i^t-\bm{p}_{i, \text{landmark, second}}|\le \epsilon_2. 
\end{aligned}
\end{equation}
Similarly, the STL specifications for simple spread II are:
\vspace{-5pt} % Reduce space
\begin{equation}
\begin{aligned}
    \varphi_{i_1} & =\Diamond_{[0,T]} \bigwedge_{1,2, \dots} \min_{j \in N} |\bm{p}_j^t-\bm{p}_{i, \text{landmark}, \text{first}}|\le \epsilon_1, \\
    \varphi_{i_2} &=\Diamond \Box_{[0, T]}  \bigwedge_{1,2, \dots} \min_{j \in N} |\bm{p}_j^t-\bm{p}_{i, \text{landmark}, \text{second}}|\le \epsilon_2.
\end{aligned}
\end{equation}

The common safety requirement is: $\varphi_{i_3} =\Box_{[0,T]} \bigwedge_{1,2, \dots} |\bm{p}_i^t-\bm{p}_j^t|\ge D_{safe}, \forall j \in N $. 

\begin{figure}[h]
\vspace*{-10pt}
\centering 
\subfloat[simple coordination II]{\includegraphics[height=1.1in]{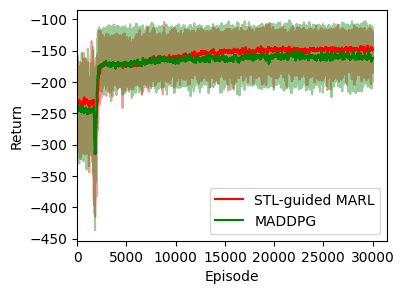}}
\subfloat[simple spread II]{\includegraphics[height=1.1in]{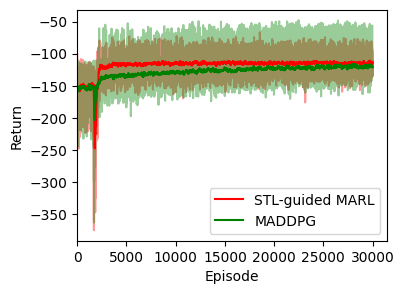}}
\caption{Training results in MPE.} \label{fig:MPE}
\vspace*{-10pt}
\end{figure}
\begin{table}[h!]
\vspace*{-5pt}
   \caption{Mean episode return in MPE. STL-guided MARL shows higher mean episode return in both tasks compared with baseline algorithm, demonstrating the advantages of our method in helping agent learn a better policy to fulfill the designer's intentions.}
   \label{table:MPE}
    %\small % or \footnotesize
   \centering
   \begin{tabular}{ c@{}c@{}c }
     \hline
     Methods & Simple coordination II  & Simple spread II \\ \hline
     STL-guided MARL &  $ \bm{-162.01} \pm 29.93$ &  $ \bm{-120.47} \pm 18.13 $ \\\hline
    MADDPG    &  $ -169.17\pm 33.64$ & $-128.36 \pm 17.90$ \\ \hline
   \end{tabular}
   \vspace*{-5pt}
\end{table}

\begin{figure}[h]
\vspace*{-10pt}
\centering 
\includegraphics[height=1.08in]{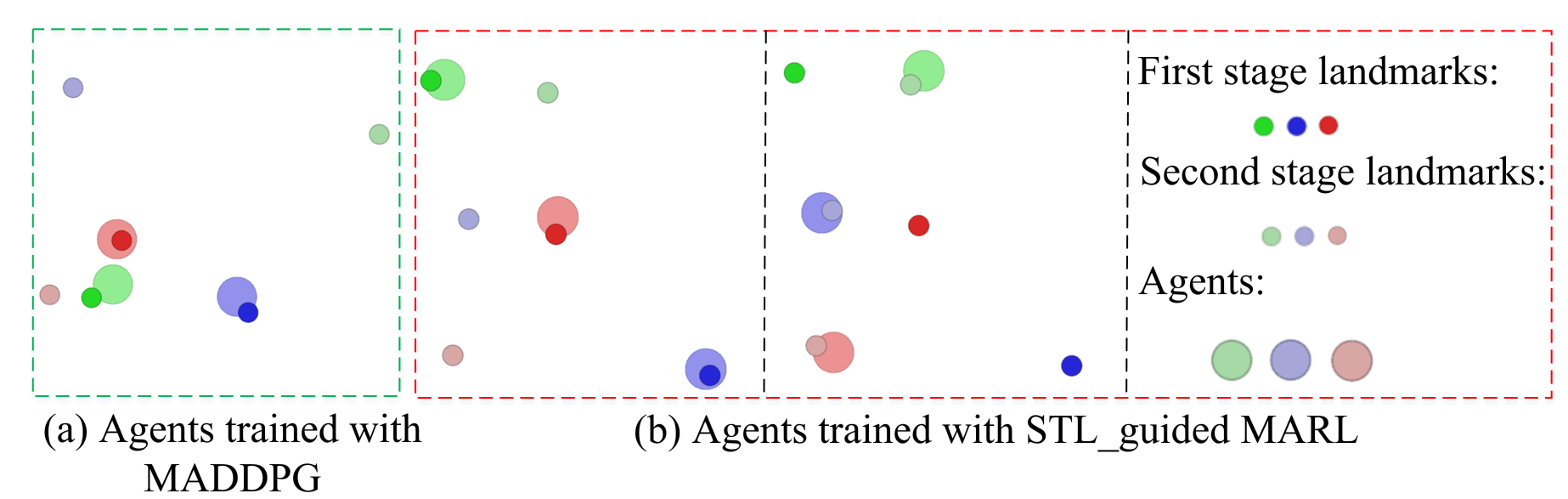}
%\subfloat[]{\includegraphics[height=0.85in]{figures/laststage2.png}}
\caption{Simple coordination II in MPE. Agents trained with MADDPG hover around the first stage landmarks and fail to cover the second stage as shown in (a). Agents trained with STL-guided MARL visit the first stage landmarks and then covers the second stage landmarks as shown in (b).} \label{fig:MPE coordination II}
\vspace*{-20pt}
\end{figure}

\paragraph{Experiment Results}
We train the algorithms for 30000 episodes, with episode length of 25 to evaluate their performance. Fig.~\ref{fig:MPE} provides insights into the mean and variance of the average returns for the two tasks. Notably, the STL-guided MARL algorithm demonstrates superior performance compared to baseline algorithm in terms of episode return. Specifically, as shown in Fig.~\ref{fig:MPE coordination II}, the agents trained with the STL-guided MARL approach exhibit a remarkable ability to cover the second stage landmarks after visiting the first stage. On the other hand, the agents trained with baseline algorithm MADDPG struggle to cover the second stage landmarks and tend to hover around the first stage.

This disparity in performance highlights the advantage of the STL-guided approach in facilitating the policy learning. By incorporating STL specifications, the agents are encouraged to adhere to specific behavioral patterns that result in more successful navigation and completion of the tasks. In contrast, the agents trained with a comparison reward, without the STL-guided framework, lack the guidance necessary to achieve optimal performance and struggle to exhibit the desired behavior.
\iffalse
\begin{table}[h!]
    \caption{Common hyperparameters for STL-guided MARL, MADDPG in MPE testbed}
    \label{table:hyperparameters mpe}
    \centering
    \begin{tabular}{ll}
        \hline
        Hyperparameter & Value \\
        \hline 
        actor learning rate & 1e-3 \\
        critic learning  & 1e-3 \\
        critic loss & mse loss \\
        discount Factor & 0.95 \\
        batch size & 1024 \\
        buffer capacity & 1e6 \\
        optimizer & Adam \\
        fc layer dim & 64 \\
        activation layer & ReLU \\
        \hline
    \end{tabular}
\end{table}
\fi

\begin{table*}[t]
    \caption{Mean episode return and safety rate in \textit{Traffic-jam} scenario.}
   \label{table:traffic-jam}
 % \small % or \footnotesize
   \centering
   \begin{tabularx}{\textwidth}{@{}c  X  X  X c @{}}
     \hline
     \multirow{2}{*}{Methods} &\multicolumn{3}{c}{Mean episode return} & \multirow{2}{*}{Safety rate} \\
     \cline{2-4}
     & \ \ \ \ Agent 1 & \ \ \ \ Agent 2 & \ \ \ \ Agent 3 &  \\
     \hline
     STL-guided MARL
       & $\bm{1469.98} \pm 43.87 $ &  $\bm{3577.53} \pm 580.04 $ & $\bm{3699.26} \pm 588.62 $  & $ \bm{97}\%$ \\
     \hline
    MAPPO
     & $ 1244.12\pm 174.18 $  & $1838.96 \pm 392.22 $ &  $2245.07\pm 150.75$ & $74\%$ \\
     \hline
    MAA2C
      & $1131.51 \pm 389.62 $ &  $ 1645.80 \pm 693.49 $ &  $2713.19 \pm 465.45$ & $88\%$ \\
     \hline
    MAPPO w/o STL safety shield
      & $853.30 \pm 421.46 $ &  $1241.91 \pm 223.38 $ &  $2102.25 \pm 713.09$ & $65\%$ \\
      \hline
   \end{tabularx}
 \vspace*{-10pt}
\end{table*}

\subsection{CARLA Testbed}
\label{section: carla}
The \textit{Traffic-jam} scenario settings and STL specifications are illustrated in Section~\ref{case_study}. To further validate our algorithm, we add 3 more autonomous vehicles (agents) in the \textit{Traffic-jam} scenario, we name it as \textit{Traffic-jam Expansion}. 
\paragraph{Baselines}
We adopt the reward that are widely used in the existing literature for lane merging case\cite{nakka2022multi, chen2021deep}. The reward for agent $i$ are defined as follow: $r_i = w_1 r_i^{\text{speed}} + w_2 r_i^{\text{collision}} + w_3 r_i^{\text{dest}} $, where $w_1, w_2, w_3 \in \mathbb{R}$ are the weights, $r_i^{\text{speed}} = \frac{|\bm{v}_i|}{v_{max}}$, $r_i^{\text{collision}} = -I_{\text{col}}$ with $I_{\text{col}}$ being the collision intensity collected by collision sensor, $r_i^{\text{dest}} = -|\bm{p}_i - \bm{p}_{\text{dest}}|+ c$ with $c$ being a constant. we use MAPPO algorithm \cite{yu2022surprising}, MAA2C algorithm\cite{papoudakis2020benchmarking}, and MAPPO algorithm without STL safety shield as our baseline algorithms.

\begin{figure}[h]
\centering
\subfloat[MAPPO w/o STL safety shield]{\includegraphics[height=1in]{figures/traffic_no1.png}} \ \ 
\subfloat[STL-guided MARL]{\includegraphics[height=1in]{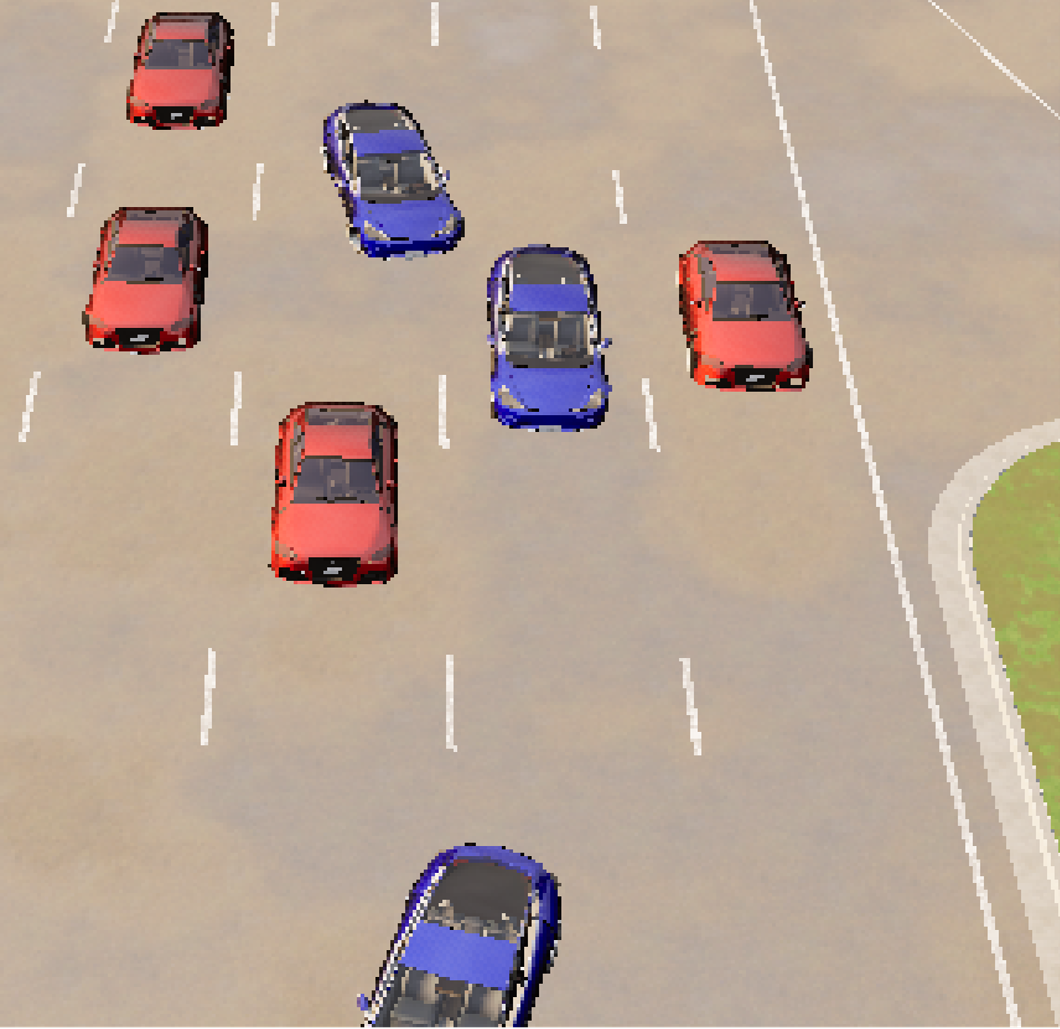}}  \ \ 
\subfloat[STL-guided MARL]{\includegraphics[height=1in]{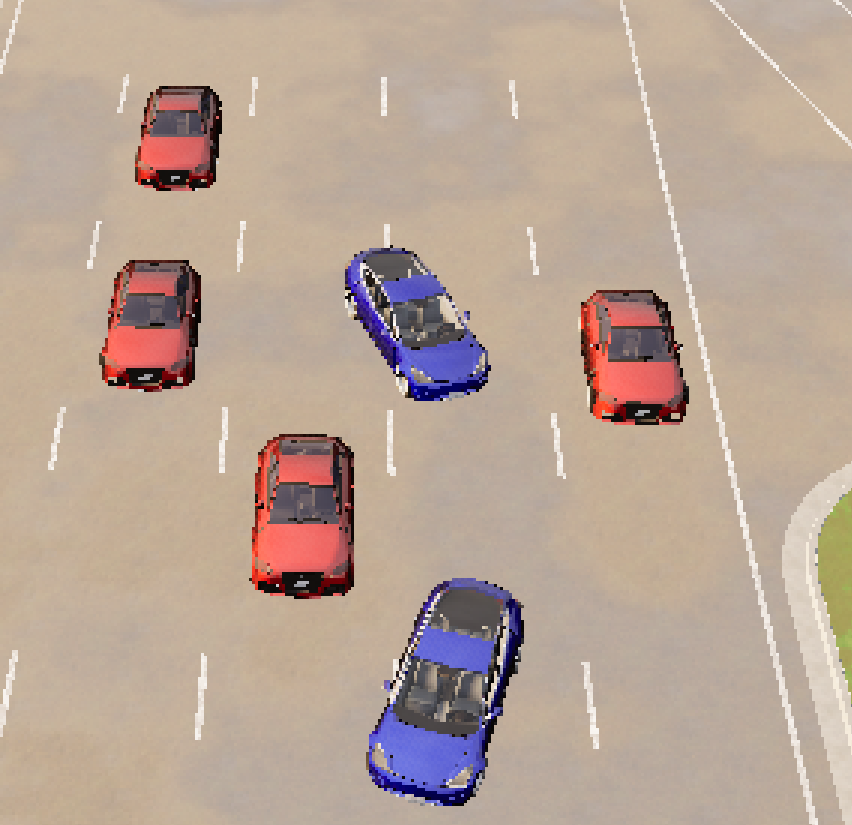}}
\caption{\textit{Traffic-jam} scenario testing process in CARLA. (a): Agents trained with the baseline algorithm fail to bypass the broken-down vehicle and collide. (b) and (c): Agents trained with STL-guided MARL successfully pass through the only open lane without collision in a timely manner.} \label{fig:Carla jam}
\vspace*{-10pt}
\end{figure}

\begin{table}[ht]
   \caption{Total mean episode return and safety rate in \textit{Traffic-jam Expansion} Scenario. "Total Mean episode return" represents the aggregated total of mean episode returns across all agents.}
   \label{table:expansion}
   \centering
   \renewcommand{\arraystretch}{1.2} % Increase line spacing within cells
   \begin{tabular}{lcc}
   \hline
      Methods & Total mean episode return & Safety rate \\ \hline
      STL-guided MARL & $\bm{31828.34} \pm 2772.05$ & $\bm{90\%}$ \\ \hline
      MAPPO & $27617.27 \pm 9504.04$ & $64\%$ \\ \hline
      MAA2C & $24169.17 \pm 4133.64$ & $68\%$ \\ \hline
      \multirow{1.2}{*}{MAPPO w/o} & \multirow{2}{*}{$21409.98 \pm 2848.80$} & \multirow{2}{*}{$20\%$} \\
       STL safety shield & & \\ \hline
   \end{tabular}
   \vspace*{-20pt}
\end{table}

\paragraph{Experiment Results}

All the algorithms are trained 100 episodes with the episode length of 150 steps. For \textit{Traffic-jam} scenario, training results are shown in Table~\ref{table:traffic-jam} and the testing visualization is shown in Fig.~\ref{fig:Carla jam}. For \textit{Traffic-jam Expansion} scenario, training results are shown in Table~\ref{table:expansion}. Here safety rate is defined as the proportion of episodes with no collisions relative to the total number of episodes. It can be observed that:
(1) STL-guided MARL with STL safety shield largely outperforms the algorithm without it in terms of safety rate, therefore, showing the effectiveness of our proposed STL safety shield in ensuring the safety of the system.
%Note that the reason of $3\%$ safety loss might be the models mismatch, i.e., our identified system model is not exactly the same with the real model run in the simulator, which is hard to obtain.
(2) While the STL specifications for each agent remain the same, the mode of interaction during an episode can vary between competitive and collaborative. Our proposed method consistently outperforms the baseline methods in terms of the total mean episode return, as demonstrated in Table~\ref{table:expansion}.  This demonstrates our algorithm can work in mixed cooperative-competitive environments. 
(3) The STL-guided MARL algorithm consistently outperforms the baseline algorithms in mean episode return. The superior performance of the STL-guided MARL algorithm can be attributed to its expressiveness and ability to capture the designer's goal. By leveraging STL  as a guidance framework, the algorithm is able to incorporate high-level specifications and constraints into the learning process. This enables the agent to learn a policy that aligns more closely with the desired behavior outlined by the designer. Thus, the STL-guided MARL algorithm demonstrates its effectiveness in improving the learning process and enabling the agent to achieve better performance.

\section{Conclusion}
%In order to tackle the challenge of reward designing in multi-agent reinforcement learning, We propose an STL guided multi-agent learning algorithm to help the agent learn to fulfill the designer's intentions. The algorithm checks the partial trajectory during training and provide the robustness value as the reward given STL formula. Moreover, the safety requirements specified by STL formulas are used to guide the CBF layer to safeguard the actions chosen by MARL. Though empirical study in MPE and CARLA testbeds, we demonstrate the advantages of our proposed algorithm in helping the agents learn better policies and improving the safety of the system.

We propose a multi-agent reinforcement learning algorithm that leverages signal temporal logic (STL) specifications to guide the learning process and ensure the satisfaction of safety requirements and task objectives for each agent. By incorporating STL safety shields, our algorithm provides additional safety guarantees in the system. Through case studies, we demonstrate that our approach outperforms traditional MARL methods with hand-engineered rewards, as it learns better policies with higher average rewards and ensures the system safety. Our work highlights the potential of using temporal logic and formal languages in MARL to address the challenges of reward design and safety in complex multi-agent systems.

\bibliographystyle{unsrt}
\bibliography{reference}

\begin{thebibliography}{10}

\bibitem{silver2021reward}
David Silver, Satinder Singh, Doina Precup, and Richard~S Sutton.
\newblock Reward is enough.
\newblock {\em Artificial Intelligence}, 299:103535, 2021.

\bibitem{lu2021decentralized}
Songtao Lu, Kaiqing Zhang, Tianyi Chen, Tamer Ba{\c{s}}ar, and Lior Horesh.
\newblock Decentralized policy gradient descent ascent for safe multi-agent reinforcement learning.
\newblock In {\em Proceedings of the AAAI Conference on Artificial Intelligence}, volume~35, pages 8767--8775, 2021.

\bibitem{zhang2022spatial}
Zhili Zhang, Songyang Han, Jiangwei Wang, and Fei Miao.
\newblock Spatial-temporal-aware safe multi-agent reinforcement learning of connected autonomous vehicles in challenging scenarios.
\newblock {\em arXiv preprint arXiv:2210.02300}, 2022.

\bibitem{garcia2015comprehensive}
Javier Garc{\i}a and Fernando Fern{\'a}ndez.
\newblock A comprehensive survey on safe reinforcement learning.
\newblock {\em Journal of Machine Learning Research}, 16(1):1437--1480, 2015.

\bibitem{brunke2022safe}
Lukas Brunke, Melissa Greeff, Adam~W Hall, Zhaocong Yuan, Siqi Zhou, Jacopo Panerati, and Angela~P Schoellig.
\newblock Safe learning in robotics: From learning-based control to safe reinforcement learning.
\newblock {\em Annual Review of Control, Robotics, and Autonomous Systems}, 5:411--444, 2022.

\bibitem{zhao2023state}
Weiye Zhao, Tairan He, Rui Chen, Tianhao Wei, and Changliu Liu.
\newblock State-wise safe reinforcement learning: A survey.
\newblock {\em arXiv preprint arXiv:2302.03122}, 2023.

\bibitem{corazza2022reinforcement}
Jan Corazza, Ivan Gavran, and Daniel Neider.
\newblock Reinforcement learning with stochastic reward machines.
\newblock In {\em Proceedings of the AAAI Conference on Artificial Intelligence}, volume~36, pages 6429--6436, 2022.

\bibitem{kantaros2022accelerated}
Yiannis Kantaros.
\newblock Accelerated reinforcement learning for temporal logic control objectives.
\newblock In {\em 2022 IEEE/RSJ International Conference on Intelligent Robots and Systems (IROS)}, pages 5077--5082. IEEE, 2022.

\bibitem{balakrishnan2019structured}
Anand Balakrishnan and Jyotirmoy~V Deshmukh.
\newblock Structured reward shaping using signal temporal logic specifications.
\newblock In {\em 2019 IEEE/RSJ International Conference on Intelligent Robots and Systems (IROS)}, pages 3481--3486. IEEE, 2019.

\bibitem{hammond2021multi}
Lewis Hammond, Alessandro Abate, Julian Gutierrez, and Michael Wooldridge.
\newblock Multi-agent reinforcement learning with temporal logic specifications.
\newblock {\em arXiv preprint arXiv:2102.00582}, 2021.

\bibitem{zhang2021multi}
Kaiqing Zhang, Zhuoran Yang, and Tamer Ba{\c{s}}ar.
\newblock Multi-agent reinforcement learning: A selective overview of theories and algorithms.
\newblock {\em Handbook of reinforcement learning and control}, pages 321--384, 2021.

\bibitem{cui2019multi}
Jingjing Cui, Yuanwei Liu, and Arumugam Nallanathan.
\newblock Multi-agent reinforcement learning-based resource allocation for uav networks.
\newblock {\em IEEE Transactions on Wireless Communications}, 19(2):729--743, 2019.

\bibitem{qie2019joint}
Han Qie, Dianxi Shi, Tianlong Shen, Xinhai Xu, Yuan Li, and Liujing Wang.
\newblock Joint optimization of multi-uav target assignment and path planning based on multi-agent reinforcement learning.
\newblock {\em IEEE access}, 7:146264--146272, 2019.

\bibitem{chu2019multi}
Tianshu Chu, Jie Wang, Lara Codec{\`a}, and Zhaojian Li.
\newblock Multi-agent deep reinforcement learning for large-scale traffic signal control.
\newblock {\em IEEE Transactions on Intelligent Transportation Systems}, 21(3):1086--1095, 2019.

\bibitem{shalev2016safe}
Shai Shalev-Shwartz, Shaked Shammah, and Amnon Shashua.
\newblock Safe, multi-agent, reinforcement learning for autonomous driving.
\newblock {\em arXiv preprint arXiv:1610.03295}, 2016.

\bibitem{amodei2016concrete}
Dario Amodei, Chris Olah, Jacob Steinhardt, Paul Christiano, John Schulman, and Dan Man{\'e}.
\newblock Concrete problems in ai safety.
\newblock {\em arXiv preprint arXiv:1606.06565}, 2016.

\bibitem{proper2012modeling}
Scott Proper and Kagan Tumer.
\newblock Modeling difference rewards for multiagent learning.
\newblock In {\em AAMAS}, pages 1397--1398, 2012.

\bibitem{icarte2018using}
Rodrigo~Toro Icarte, Toryn Klassen, Richard Valenzano, and Sheila McIlraith.
\newblock Using reward machines for high-level task specification and decomposition in reinforcement learning.
\newblock In {\em International Conference on Machine Learning}, pages 2107--2116. PMLR, 2018.

\bibitem{li2019formal}
Xiao Li, Zachary Serlin, Guang Yang, and Calin Belta.
\newblock A formal methods approach to interpretable reinforcement learning for robotic planning.
\newblock {\em Science Robotics}, 4(37):eaay6276, 2019.

\bibitem{cai2023safe}
Mingyu Cai, Shaoping Xiao, Junchao Li, and Zhen Kan.
\newblock Safe reinforcement learning under temporal logic with reward design and quantum action selection.
\newblock {\em Scientific reports}, 13(1):1925, 2023.

\bibitem{li2017reinforcement}
Xiao Li, Cristian-Ioan Vasile, and Calin Belta.
\newblock Reinforcement learning with temporal logic rewards.
\newblock In {\em 2017 IEEE/RSJ International Conference on Intelligent Robots and Systems (IROS)}, pages 3834--3839. IEEE, 2017.

\bibitem{elsayed2021safe}
Ingy ElSayed-Aly, Suda Bharadwaj, Christopher Amato, R{\"u}diger Ehlers, Ufuk Topcu, and Lu~Feng.
\newblock Safe multi-agent reinforcement learning via shielding.
\newblock {\em arXiv preprint arXiv:2101.11196}, 2021.

\bibitem{leon2020extended}
Borja~G Le{\'o}n and Francesco Belardinelli.
\newblock Extended markov games to learn multiple tasks in multi-agent reinforcement learning.
\newblock {\em arXiv preprint arXiv:2002.06000}, 2020.

\bibitem{maler2004monitoring}
Oded Maler and Dejan Nickovic.
\newblock Monitoring temporal properties of continuous signals.
\newblock In {\em Formal Techniques, Modelling and Analysis of Timed and Fault-Tolerant Systems}, pages 152--166. Springer, 2004.

\bibitem{fainekos2009robustness}
Georgios~E Fainekos and George~J Pappas.
\newblock Robustness of temporal logic specifications for continuous-time signals.
\newblock {\em Theoretical Computer Science}, 410(42):4262--4291, 2009.

\bibitem{deshmukh2017robust}
Jyotirmoy~V Deshmukh, Alexandre Donz{\'e}, Shromona Ghosh, Xiaoqing Jin, Garvit Juniwal, and Sanjit~A Seshia.
\newblock Robust online monitoring of signal temporal logic.
\newblock {\em Formal Methods in System Design}, 51(1):5--30, 2017.

\bibitem{littman1994markov}
Michael~L Littman.
\newblock Markov games as a framework for multi-agent reinforcement learning.
\newblock In {\em Machine learning proceedings 1994}, pages 157--163. Elsevier, 1994.

\bibitem{lowe2017multi}
Ryan Lowe, Yi~I Wu, Aviv Tamar, Jean Harb, OpenAI Pieter~Abbeel, and Igor Mordatch.
\newblock Multi-agent actor-critic for mixed cooperative-competitive environments.
\newblock {\em Advances in neural information processing systems}, 30, 2017.

\bibitem{yu2022surprising}
Chao Yu, Akash Velu, Eugene Vinitsky, Jiaxuan Gao, Yu~Wang, Alexandre Bayen, and Yi~Wu.
\newblock The surprising effectiveness of ppo in cooperative multi-agent games.
\newblock {\em Advances in Neural Information Processing Systems}, 35:24611--24624, 2022.

\bibitem{mnih2016asynchronous}
Volodymyr Mnih, Adria~Puigdomenech Badia, Mehdi Mirza, Alex Graves, Timothy Lillicrap, Tim Harley, David Silver, and Koray Kavukcuoglu.
\newblock Asynchronous methods for deep reinforcement learning.
\newblock In {\em International conference on machine learning}, pages 1928--1937. PMLR, 2016.

\bibitem{schulman2015high}
John Schulman, Philipp Moritz, Sergey Levine, Michael Jordan, and Pieter Abbeel.
\newblock High-dimensional continuous control using generalized advantage estimation.
\newblock {\em arXiv preprint arXiv:1506.02438}, 2015.

\bibitem{lindemann2018control}
Lars Lindemann and Dimos~V Dimarogonas.
\newblock Control barrier functions for signal temporal logic tasks.
\newblock {\em IEEE control systems letters}, 3(1):96--101, 2018.

\bibitem{kong2015autonomous}
J~Kong, M~Pfeiffer, G~Schildbach, and F~Borrelli.
\newblock Autonomous driving using model predictive control and a kinematic bicycle vehicle model.
\newblock In {\em Intelligent Vehicles Symposium, Seoul, Korea}, 2015.

\bibitem{dixit2018trajectory}
Shilp Dixit, Saber Fallah, Umberto Montanaro, Mehrdad Dianati, Alan Stevens, Francis Mccullough, and Alexandros Mouzakitis.
\newblock Trajectory planning and tracking for autonomous overtaking: State-of-the-art and future prospects.
\newblock {\em Annual Reviews in Control}, 45:76--86, 2018.

\bibitem{cesari2017scenario}
Gianluca Cesari, Georg Schildbach, Ashwin Carvalho, and Francesco Borrelli.
\newblock Scenario model predictive control for lane change assistance and autonomous driving on highways.
\newblock {\em IEEE Intelligent transportation systems magazine}, 9(3):23--35, 2017.

\bibitem{ames2016control}
Aaron~D Ames, Xiangru Xu, Jessy~W Grizzle, and Paulo Tabuada.
\newblock Control barrier function based quadratic programs for safety critical systems.
\newblock {\em IEEE Transactions on Automatic Control}, 62(8):3861--3876, 2016.

\bibitem{zeng2021safety}
Jun Zeng, Bike Zhang, and Koushil Sreenath.
\newblock Safety-critical model predictive control with discrete-time control barrier function.
\newblock In {\em 2021 American Control Conference (ACC)}, pages 3882--3889. IEEE, 2021.

\bibitem{dosovitskiy2017carla}
Alexey Dosovitskiy, German Ros, Felipe Codevilla, Antonio Lopez, and Vladlen Koltun.
\newblock Carla: An open urban driving simulator.
\newblock In {\em Conference on robot learning}, pages 1--16. PMLR, 2017.

\bibitem{nakka2022multi}
Sai Krishna~Sumanth Nakka, Behdad Chalaki, and Andreas~A Malikopoulos.
\newblock A multi-agent deep reinforcement learning coordination framework for connected and automated vehicles at merging roadways.
\newblock In {\em 2022 American Control Conference (ACC)}, pages 3297--3302. IEEE, 2022.

\bibitem{chen2021deep}
Dong Chen, Mohammad Hajidavalloo, Zhaojian Li, Kaian Chen, Yongqiang Wang, Longsheng Jiang, and Yue Wang.
\newblock Deep multi-agent reinforcement learning for highway on-ramp merging in mixed traffic.
\newblock {\em arXiv preprint arXiv:2105.05701}, 2021.

\bibitem{papoudakis2020benchmarking}
Georgios Papoudakis, Filippos Christianos, Lukas Sch{\"a}fer, and Stefano~V Albrecht.
\newblock Benchmarking multi-agent deep reinforcement learning algorithms in cooperative tasks.
\newblock {\em arXiv preprint arXiv:2006.07869}, 2020.

\end{thebibliography}

\end{document}